\documentclass[10pt,twocolumn,letterpaper]{article}

\usepackage{cvpr}
\usepackage{times}
\usepackage{epsfig}
\usepackage{graphicx}
\usepackage{amsmath}
\usepackage{amssymb}
\usepackage{lipsum}
\usepackage{tabls}
\usepackage{pbox}
\usepackage{array}
\usepackage{cuted}
\usepackage[pagebackref=true,breaklinks=true,letterpaper=true,colorlinks,bookmarks=false]{hyperref}
\usepackage{cleveref}

\newcommand{\myparagraph}[1]{\vspace{0.5em}\noindent {\bf #1.}}
\newcommand{\bx}{\mathbf{x}}

\cvprfinalcopy

\title{Self-supervised Learning of Geometrically Stable Features Through Probabilistic Introspection}

\author{
David Novotny$^{1,2,}$\thanks{Authors contributed equally.} ~ ~ Samuel Albanie$^{1,}$\footnotemark[1] ~~ Diane Larlus$^2$ ~ ~ Andrea Vedaldi$^1$ \\
\begin{minipage}{.4\textwidth}
\centering
$^1$\small{Visual Geometry Group\\Dept. of Engineering Science, University of Oxford\\}
{\tt\small \{david,albanie,vedaldi\}@robots.ox.ac.uk}
\end{minipage} 
\begin{minipage}{.4\textwidth}
\centering
$^2$\small{Computer Vision Group\\NAVER LABS Europe\\} 
{\tt\small diane.larlus@naverlabs.com}
\end{minipage}
}

\begin{document}
\maketitle

\begin{abstract}
Self-supervision can dramatically cut back the amount of manually-labelled data required to train deep neural networks. While self-supervision has usually been considered for tasks such as image classification, in this paper we aim at extending it to geometry-oriented tasks such as semantic matching and part detection. We do so by building on several recent ideas in unsupervised landmark detection. Our approach learns dense distinctive visual descriptors from an unlabeled dataset of images using synthetic image transformations. It does so by means of a robust probabilistic formulation that can introspectively determine which image regions are likely to result in stable image matching. We show empirically that a network pre-trained in this manner requires significantly less supervision to learn semantic object parts compared to numerous pre-training alternatives. We also show that the pre-trained representation is excellent for semantic object matching.
\end{abstract}
\section{Introduction}\label{s:intro}

One factor that limits the applicability of deep neural networks to many practical problems is the cost of procuring a sufficient amount of  supervised data for learning. This explains the increasing interest in techniques that can learn good deep representations \emph{without the use of manual supervision}. Methods that rely on self-supervision~\cite{doersch2015unsupervised,noroozi2016,pathakCVPR16context}, in particular, can initialize deep neural networks from unlabeled image collections. The resulting pre-trained networks can then be fine-tuned to solve a desired task with far fewer manual annotations than would be required if they were trained from scratch.

\begin{figure}[t]
\centering
\includegraphics[width=\linewidth]{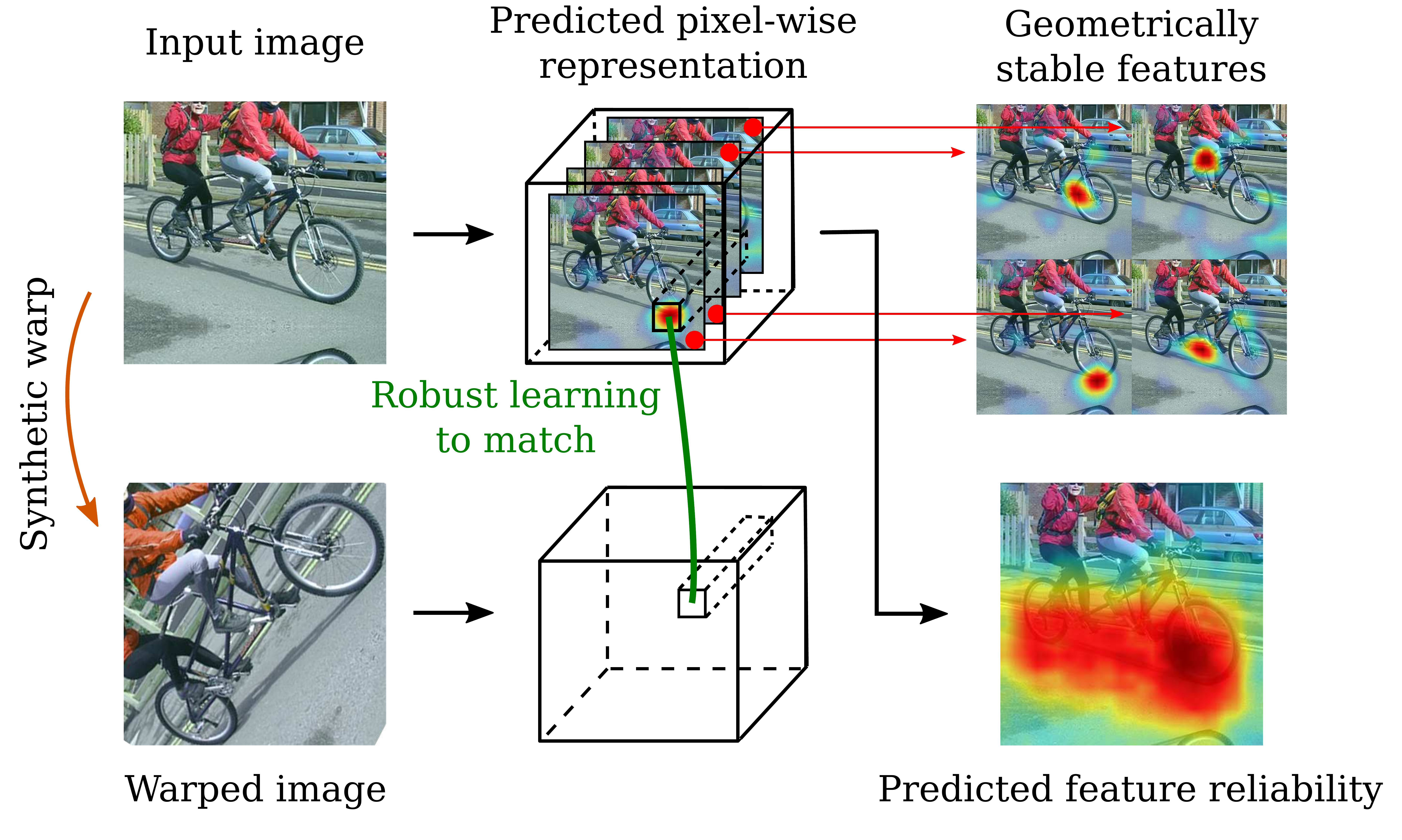}
\caption{Our approach leverages correspondences obtained from synthetic warps in order to self-supervise the learning of a dense image representation. This results in highly localized and geometrically stable features. The use of a novel robust probabilistic formulation allows to additionally predict a pixel-level confidence map that estimates the matching ability of these features.}\label{f:spalsh}
\end{figure}

While several authors have looked at self-supervision for tasks such as image classification and segmentation, less work has been done on tasks that involve understanding the geometric properties of object categories. In this paper, therefore, we propose a self-supervised pre-training technique that obtains image representations suitable for geometry-oriented tasks. We consider two representative problems: semantic part detection and semantic matching, both of which help to characterize the geometric structure of objects.

Our specific goal is to pre-train convolutional neural networks suitable for such geometry-oriented tasks given only a dataset of images of one or more object categories \emph{with no bounding box, part or other types of geometric annotations}. Our approach is based on three ideas. First, we configure the network to compute a dense field of visual descriptors. These descriptors are learned to match corresponding object points in different images using a pairwise loss formulation. However, since no labels are given, correspondences between images are unknown. Thus, the second idea is to generate image pairs for which correspondences are known by means of \emph{synthetic warps}~\cite{kanazawa16warpnet,rocco17convolutional,thewlis17unsupervised,thewlis17Bunsupervised}. Learning from this data results in visual descriptors that are invariant to image deformations, but that may not be consistent across intra-class variations. The authors of~\cite{thewlis17Bunsupervised} suggest that intra-class generalization can be achieved by limiting the descriptor dimensionality. However, we found this approach to be too fragile to handle complex 3D object categories, particularly when many landmarks can be occluded in different views. This contrasts with other recent approaches such as AnchorNet~\cite{novotny17anchornet}, which can learn landmarks more robustly, albeit with reduced geometric accuracy.

Seeking to retain the robustness of methods such as AnchorNet~\cite{novotny17anchornet} while incorporating a geometric prior such as~\cite{thewlis17Bunsupervised}, we propose to trade-off robustness for a higher dimensionality of the descriptors. We further improve robustness by casting learning into a probabilistic formulation, our third idea. This formulation allows the network to explicitly learn, along with the visual descriptors, an estimate of their expected matching reliability.  In this manner, the network learns failure modalities, such as extracting descriptors in correspondence of background regions instead of the object or occlusions.

The resulting formulation is able to pre-train excellent networks for semantic matching and semantic part detection. This is demonstrated empirically by means of thorough experiments against a range of baselines on standard benchmark datasets. For semantic matching, our results outperform~\cite{novotny17anchornet} and~\cite{thewlis17Bunsupervised} that use a comparable level of supervision and are on par with the fully supervised method of~\cite{han2017scnet}. For part detection, we consider a few-shot keypoint detection task and show that our method performs better than all competitors when few annotations are available.

The rest of the manuscript is organized as follows. \Cref{s:related} discusses  related work, \cref{s:method} presents the technical details of our method, \cref{s:exp} conducts the experimental evaluation, and \cref{s:conc} summarizes our findings.

\section{Related Work}\label{s:related}

\myparagraph{Learning features for geometric tasks}
Hand-crafted features such as SIFT \cite{lowe2004}, DAISY \cite{yang2014daisyfilterflow}, or HOG \cite{dalal05hog},
initially designed for geometrical tasks such as matching-based retrieval \cite{Sivic2003}, stereo matching \cite{okutomi1993}, or optical flow \cite{horn93determining} formed the gold standard until very recently due to their appealing properties such as repeatability.

Dense semantic matching methods, pioneered by SIFT Flow~\cite{liu11siftflow} are designed to deal with more variability in appearance and create dense correspondences across different scenes. Following the success of CNN architectures for recognition tasks like image classification \cite{krizhevsky12imagenet}, these architectures have been used as feature extractors for other tasks, including semantic matching. Yet, without any further training, they have been shown not to improve over hand-engineered features for geometric tasks \cite{long2014do,ham2016} and most approaches still combine hand-crafted features and spatial regularization \cite{bristow2015dense,hur15generalized,kim2013deformable,liu11siftflow,zhou15flowweb}.
To overcome this, deep features have been retrained for geometric tasks
~\cite{choy16universal,zhou15flowweb,han2017scnet}.
Choy \etal \cite{choy16universal} combine a fully convolutional architecture with a contrastive loss and train with a large number of annotations.  Zhou \etal \cite{zhou2016learning} require 3D models to link correspondences between images and rendered views. Han \etal \cite{han2017scnet} follow Proposal Flow \cite{ham2016} and replace the hand-crafted features with features trained end-to-end with a large amount of annotations.

 Training geometry-aware features without costly annotations such as keypoints or 3D models has only been seldomly studied \cite{novotny17anchornet,thewlis17unsupervised,thewlis17Bunsupervised,rocco17convolutional}. The AnchorNet approach \cite{novotny17anchornet} builds discriminative parts that match different object instances as well as different object categories using only image-level supervision. Other methods have proposed to replace costly manual annotations by synthetically generating image pairs \cite{thewlis17unsupervised,thewlis17Bunsupervised,rocco17convolutional}. Thewlis \etal \cite{thewlis17unsupervised} show that placing constraints on matching builds object landmarks that are not only consistently detected across the deformation of a current instance, but also across instances.
 This work was extended to a dense formulation~\cite{thewlis17Bunsupervised}, embedding objects on a sphere. Although this works well for faces, such an approach seems less appropriate for objects with a complex 3D shape.  Rocco \etal \cite{rocco17convolutional} propose a Siamese architecture for geometric matching, composed of a feature extraction part and a matching architecture that is used to predict the parameters of a synthetic transformation applied to the input image. Artificial correspondences were also used in \cite{kanazawa16warpnet} for fine-grained categories.

\myparagraph{Keypoint detection} Keypoint detection has been extremely well studied for the case of humans \cite{johnson10clusteredpose,yang11articulated,gkioxari14kposelets,andriluka142D} and recent approaches have leveraged deep architectures \cite{toshev14deeppose,tompson14joint}. Only a few works have considered keypoint detection for generic categories \cite{hejrati12analyzing3d,long2014do,xiang2014beyond,tulsiani2015viewpoints}. These methods require large training sets and none of them has considered a few-shot learning scenario.

\begin{figure*}[t!]
\includegraphics[width=\linewidth]{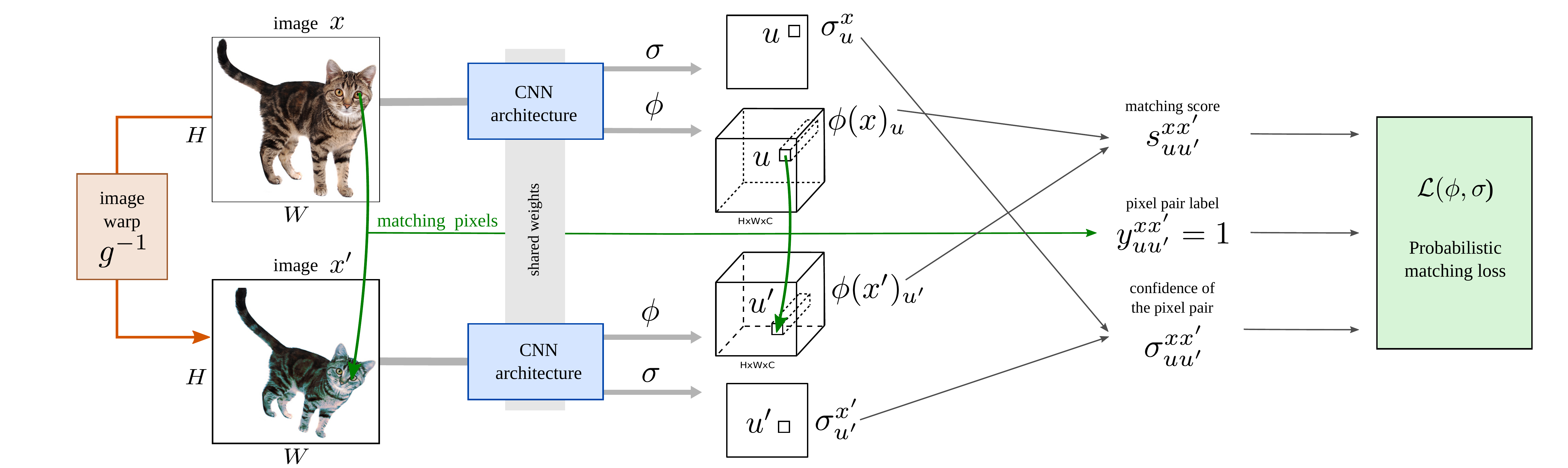}
\caption{\textbf{Overview of our approach}. Image $x$ is warped into image $x'$ using the transformation $g^{-1}$. Pairs of pixels and their labels (encoding whether they match or not according to $g^{-1}$) are used together with a probabilistic matching loss to train our architecture that predicts i) a dense image feature $\phi(x)$ and ii) a pixel level confidence value $\sigma(x)$. \label{f:arch}}
\end{figure*}

\section{Method}\label{s:method}

Our aim is to learn a neural network for object part detection and semantic matching. Furthermore, we assume that only a small number of images annotated with information relevant to these tasks is available, but that images labeled only with the presence of a given object category are plentiful. Thus, our goal is to develop a self-supervised method that can use such image-level annotations to pre-train a network that captures the object geometry.

Formally, let $\mathcal{X} = \{\bx_1,\dots,\bx_N\}$ be a collection of $N$ unlabeled images $\bx_i \in \mathbb{R}^{H\times W \times 3}$ of one or more object categories and let $\phi:\mathbb{R}^{H\times W\times 3}\rightarrow\mathbb{R}^{H\times W\times C}$ be a deep neural network extracting a dense set of feature vectors from the image. We will use the symbol $\phi(\bx)_u \in \mathbb{R}^C$ to denote the feature vector extracted at location\footnote{In our implementation, features are extracted at a lower resolution than the input image, but for clarity we ignore this difference in the notation.} \mbox{$u \in \{1,\dots,H\}\times\{1,\dots,W\}$}, namely:
$$
\forall c\in\{1,\dots,C\}:
\quad [\phi(\bx)_u]_c = [\phi(\bx)]_{uc}.
$$
Each vector $\phi(\bx)_u$ can be thought of as a descriptor of the image appearance around location $u$. Since our aim is to recognize and match object parts, we would like such descriptors to be \emph{characteristic of specific object landmarks}.

In a supervised setting, one is given the identity of the object part found at each location $u$ and can use this information to learn the descriptors. However, in our case this information is \emph{not} available, so we must resort to a different supervisory signal. We do so by constraining descriptors to be invariant (\cref{s:equivariance}) and discriminative (\cref{s:discrim}) with respect to synthetic image transformations, and make this robust using a form of probabilistic introspection (\cref{s:prob}). The resulting learning objective is given in \cref{s:obj} and further discussed in~\cref{s:disc}. \Cref{f:arch} provides an overview of the overall approach.

\subsection{Invariant description}\label{s:equivariance}

We say that locations $u$ and $u'$ in image $\bx$ and $\bx'$ \emph{correspond} if they are projection of the same 3D object point. For object categories, we define correspondences by analogy (such as being centered on the right eyes of two animals).

The \emph{invariance} condition states that the descriptors computed at corresponding image locations $u$ and $u'$ should be identical:
\begin{equation}\label{e:inv}
  \phi(\bx)_u = \phi(\bx')_{u'}
\end{equation}

While correspondences are not known for arbitrary images in the database $\mathcal{X}$ (short of providing manual annotations), we can at least \emph{synthetically generate} such examples. To this end, let $g : \mathbb{R}^2 \rightarrow \mathbb{R}^2, u \mapsto u'=g(u)$ be a random image warp and let $\bx' = \bx \circ g^{-1}$ be the image obtained by warping $\bx\in\mathcal{X}$ accordingly.\footnote{Here $\bx'$ is obtained from $\bx$ using inverse warp.} Then, constraint~\eqref{e:inv} can be rewritten as:
\begin{equation}\label{e:synthinv}
  \forall g, u:
  \quad
  \phi(\bx)_u = \phi(\bx \circ g^{-1})_{g(u)}
\end{equation}

While the network $\phi$ should satisfy constraint~\eqref{e:synthinv}, the latter is insufficient to characterize good descriptors as it can be trivially satisfied by making all descriptors identical. The missing ingredient is that the descriptors should also \emph{uniquely  identify} a specific object point. Building this additional constraint into the model is discussed in the next section.

\subsection{Informative invariant description}\label{s:discrim}

Invariance~\eqref{e:synthinv} must be paired with the fact that descriptors should be able to robustly distinguish between \emph{different} object points. To encode such a constraint, we note first that it does not make sense to check for exact descriptor equality or inequality as literally suggested by~\cref{e:synthinv}. Instead, descriptors are compared continuously by considering a \emph{matching score}. We define the latter to be their rectified inner product
\begin{equation}\label{e:mscore}
s_{u u'}^{\bx \bx'} = \text{max} \{0,~\langle\phi(\bx)_{u}, \phi(\bx')_{u'}\rangle\}.
\end{equation}
In order to guarantee that this score is maximum when a descriptor is compared to itself ($s_{u u'}^{\bx \bx'}\leq 1, s_{u u}^{\bx \bx}=1$), descriptors are $L^2$ normalized, so that
$$
  \|\phi(\bx)_u\|_2 = 1.
$$
The inner product is rectified because, while it makes sense for similar descriptors to be parallel, dissimilar descriptors should be orthogonal rather than anti-correlated.

Next, in order to encode invariance and discriminability together, we note that each pair of points $(u,u')$ may or may not represent a valid correspondence for a given image pair $(\bx,\bx')$. This is captured by a label $y_{uu'}^{\bx\bx'} \in \{-1,0,+1\}$, where $+1$ indicates a valid correspondence , $-1$ an invalid one, and $0$ a ``borderline'' case to be ignored. Given the labels (defined from the synthetic warps in~\cref{e:labdef}), one can define a \emph{matching loss} $\ell_{uu'}^{\bx\bx'}$:
\begin{equation}\label{e:matching}
\ell_{uu'}^{\bx\bx'} = 
\begin{cases} 
1-s_{u u'}^{\bx \bx'} & y_{uu'}^{\bx\bx'} = 1, \\
0                     & y_{uu'}^{\bx\bx'} = 0, \\
s_{u u'}^{\bx \bx'}   & y_{uu'}^{\bx\bx'} = -1.
\end{cases}
\end{equation}
However, $\ell$ cannot be satisfied for all possible choices of image and pixel pairs ($\bx$, $\bx'$) and ($u$, $u'$). For example, an object point may be occluded, a pixel may belong to the background, or the match may just be too difficult for the model to express adequately. This problem is addressed in the next section.

\subsection{Probabilistic introspection}\label{s:prob}

\begin{figure}[t!]
\includegraphics[width=\linewidth]{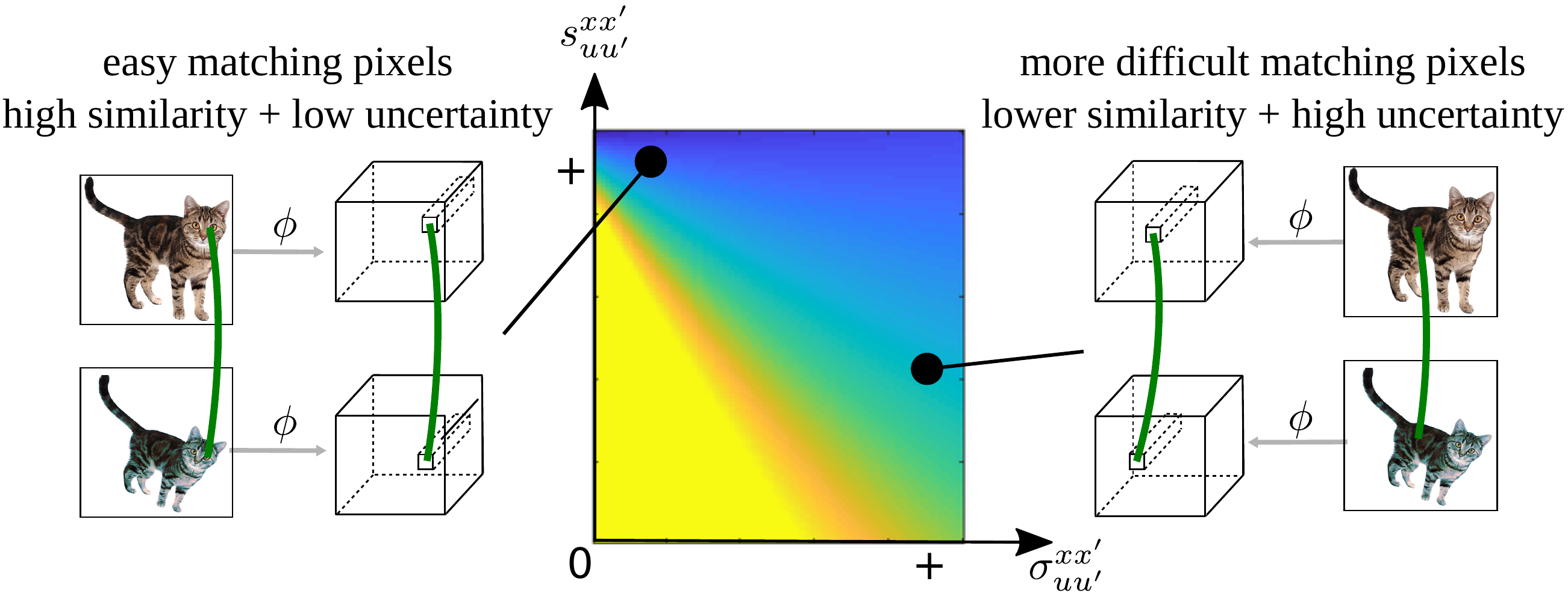}
\caption{\textbf{Illustration of the probabilistic loss.} The plot shows values of the loss for positive pairs ($y_{xx'}=1$, bluer means a smaller loss) as a function of the similarity between the pixel representations $s^{xx'}_{uu'}$ and the uncertainty $\sigma^{xx'}_{uu'}$ who's inverse $\sigma^{xx' ~ -1}_{uu'}$ corresponds to the confidence. 
The model has several options for decreasing the loss: 
(1) increasing the similarity while keeping confidence unchanged, 
(2) decreasing the confidence while keeping similarity and
(3) increasing both similarity and confidence. 
\label{f:loss}
}
\end{figure}

In order to handle difficult or impossible matches in the loss function, we do not resort to heuristics such as using robust versions of the loss~\eqref{e:matching}, but rather task the neural network with \emph{predicting when descriptors are unreliable}. In order to do so, inspired by \cite{novotny17learning,kendall2017uncertainties}, the network is modified to compute an additional scalar value $\sigma_u^\bx \in \mathbb{R^+}$ for each pixel expressing uncertainty about the quality of the descriptor extracted at $u$ and its consequent ability to establish a reliable match. Importantly, this belief is estimated from each image independently \emph{before} matching occurs. In this manner, $\sigma_u^\bx$ can be interpreted as an assessment of the informativeness of the image region that is used to compute the descriptor.

In more detail (and dropping the superscript $\bx\bx'$ for simplicity), we define a distribution over matching scores $p(s_{uu'} | y_{uu'},\sigma_{uu'})$ conditioned on the average predicted uncertainty $\sigma_{uu'} = (\sigma_u + \sigma_{u'})/2$ and on whether pixels are in correspondence or not. The distribution is given by:
\begin{equation} \label{e:score_prob}
p(s_{uu'} | y_{uu'}, \sigma_{uu'}) 
=
\frac{1}{\mathcal{C}(\sigma_{uu'})}
\exp{\frac
{1-\ell_{uu'} (s_{uu'}, y_{uu'})}
{ \sigma_{u u'}}},
\end{equation}
where $\mathcal{C}(\sigma_{uu'})$ is a normalization constant ensuring that $p(s_{uu'} | y_{uu'}, \sigma_{uu'})$ integrates to one.

To understand expression~\eqref{e:score_prob}, note that, due to the fact that $s_{uu'} \in [0,1]$ and to the particular form~\eqref{e:matching} of the function $\ell_{uu'}$ , $\mathcal{C}(\sigma_{uu'})$ is finite and does not depend on $y_{uu'}$. When the model is confident in the quality of both descriptors $\phi(\bx)_u$ and $\phi(\bx')_{u'}$, the value $\sigma_{uu'}$ is small. In this case, the distribution~\eqref{e:score_prob} has a sharp peak around 1 or 0, depending on whether pixels $(u,u')$ are in correspondence or not. On the other hand, when the model is less certain about the quality of the descriptors, the score distribution is more spread.

\subsection{Learning objective}\label{s:obj}

It is now possible to describe the overall learning objective for our method. The models $\phi$ and $\sigma$ are learned by minimizing the negative logarithm of the probability $p(s_{uu'} | y_{uu'}, \sigma_{uu'})$ averaged over images, random transformations, and point pairs. Formally, the learning objective is given by:
\begin{multline}
\mathcal{L}(\phi,\sigma) = 
\frac{1}{|\mathcal{X}|}
\sum_{\bx\in\mathcal{X}}
\mathbb{E}_{g,u,u'}
\\
\left[
- \log
p\left(
s_{uu'}^{\bx,\bx\circ g^{-1}}\!\!(\phi)
\left| y_{uu'}^g, 
\frac{\sigma_{u}^\bx + \sigma_{u'}^{\bx\circ g^{-1}}}{2}\right.\right)
\right]
\end{multline}
Here the score $s$ depends on the neural network $\phi$ as shown in~\cref{e:mscore}. The function $\sigma$ is  implemented as a small neural network branching off $\phi$ and is also learned with it. The labels $y_{uu'}^g$ are easily obtained as
\begin{equation}\label{e:labdef}
y_{uu'}^g = \begin{cases}
 1, & \|u' - g(u)\|_2 \leq \tau_1, \\
 0, & \tau_1 < \|u' - g(u)\|_2 \leq \tau_2 \\
 -1, & \text{otherwise}.
 \end{cases}  
\end{equation}
where $\tau_1<\tau_2$ are matching thresholds (we set $\tau_1=1$ and $\tau_2=30$ pixels). The value of the probabilistic loss $\mathcal{L}$ as a function of the similarity $s^{\bx\bx'}_{uu'}$ and the predicted uncertainty $\sigma_{uu'}$ is illustrated in~\Cref{f:loss}. 

The set of sampled transformations $g$ consists of random affine warps. To avoid border artifacts, following \cite{rocco17convolutional}, we mirror-pad each image enlarging its size by a factor of two while biasing the sampled transformations towards zooming into the padded image. In order to avoid  potential trivial solutions due to keeping the first image $\bx$ unwarped (as the network can catch subtle artifacts induced by warping), we sample two transformations $\hat g$, $\hat g'$ and then warp the original input image $\hat x$ twice to form the input image pair $\bx = \hat \bx \circ \hat g^{-1}$  and $\bx' = \hat \bx \circ \hat g'^{-1}$. The pairwise transformation $g = \hat g \circ \hat g'^{-1}$ \ is a straightforward composition of $\hat g$ and $\hat g'$.  In order to sample pairs of pixels $(u,u')$, we first randomly pick 700 points $\mathcal{U} = \{u_i\}_{i=1}^{700}$ from the first image. For each $u_i$, we then sample $u_i' = g(u_i)$ from the second image and evaluate the loss $\mathcal{L}$ on all possible pairs $(u_i,u_j') \in \mathcal{U} \times \mathcal{U}'$. We then follow a hard negative mining strategy by selecting the 30 negative samples $u'$  from the second image (out of 700 potential samples) that contribute to $\mathcal{L}$ the most. Backpropagation is then performed only through these ``hard negative'' examples and all the positive examples while equally balancing the overall weights of the two sets of pixel pairs.

\begin{figure}
\centering
\input{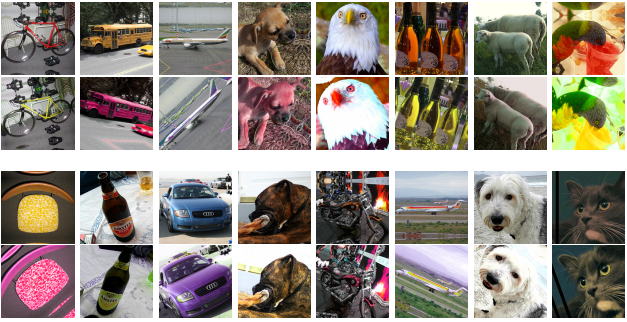}
\caption{\textbf{Example geometric and appearance transformations} used
to supervise the learning of our representation. The first (resp. third) row displays original images
while the second (resp. fourth) row shows their transformed versions.}\label{f:example-warps}
\end{figure}

\myparagraph{Appearance transformations} While random affine warping makes our features invariant to the geometric transformations, a successful representation should be also invariant to intraclass appearance variations caused by \eg color and illumination changes. Hence, besides warping the input image, we apply a random color transformation $c(\hat g(\hat \bx))$  after the geometric transformation $\hat g(\hat \bx)$. The color transformations are generated following the approach of \cite{liu2016ssd}. We increase the intensity of the color shifts in order to introduce substantial appearance changes required to boost the invariance properties of the representation. Examples of the original images and their geometry-appearance transformations are shown in \Cref{f:example-warps}.

\subsection{Discussion}\label{s:disc}

Besides its robust nature, the formulation so far can be seen as learning discriminative viewpoint invariant features. This does not guarantee  \emph{per se} that the learned descriptors are characteristics of particular object parts. For example, since the model is only trained against synthetic warps of individual images, the descriptors computed for analogous parts in different object instances (e.g.\ the eyes in two different cats) may still differ. Even out-of-plane rotations are in principle sufficient to throw off the model.

Recently, the authors of~\cite{thewlis17Bunsupervised} have suggested to constrain the descriptor capacity to favor generalization. In particular, they argue that using two dimensional descriptors strongly encourages them to attach to specific points on the surface of an object, and thus to generalize across different object instances.
Nevertheless, the method of~\cite{thewlis17Bunsupervised} was found to be too fragile to work well in challenging data where significant occlusions may be present. Our approach trades off descriptor generality for robustness. As we will see in the experiments, this pays off as, ultimately, the representation is fine-tuned with a small amount of supervised data which is sufficient to bridge most of the gaps.

\subsection{Learning details} \label{s:details}

We learn our representation using the training images of the 12 rigid PASCAL classes from the ImageNet dataset (but we test it on all 20 classes, including non-rigid ones). As a preprocessing step, we apply a weakly supervised detector \cite{bilen16} and use the resulting image crops instead of the full images. This detector only requires image-level labels and no further supervision is used. This is exactly the same level of supervision used in \cite{novotny17anchornet,rocco17convolutional} and weaker than in \cite{thewlis17unsupervised} where bounding box annotations are required.

The representation predictor $\phi(\bx)$ is a deep convolutional neural network whose architecture is based on the ResNet-50 model \cite{he2015deep} due to its good compromise between speed and capacity.  We remove the two topmost layers and base the rest of our model on the rectified res5c features. In order to increase the spatial resolution of the produced representation, following \cite{yu2015multi} we dilate all res5 convolutional filters by a factor of 2 while decreasing their stride to 1. 
Finally, we attach a $1\times1$ convolutional layer that produces raw embedding vectors $\hat \phi(\bx) \in\mathbb{R}^{H \times W \times (C+1)}$. The first $C$ channels of $\hat \phi(\bx)$ are sliced out and $\ell_2$ normalized at every spatial location $u$ to form the embedding $\phi(\bx) \in\mathbb{R}^{H \times W \times C}$. The last $(C+1)$-th channel $\phi(\bx)[:,:,C+1]$ of $\hat \phi(\bx)$ is passed through a SoftReLU and lower-bounded by $\epsilon \rightarrow 0$ which results in the inverse-confidence predictions 
\mbox{$\sigma(\bx) = \log( 1 + \exp(\hat \phi(\bx)[:,:,C+1]) ) + \epsilon$}.

Our network is optimized using the AdaGrad solver. Learning rate, weight decay and momentum  were set to $0.001$, $0.0005$ and $0.9$ respectively. 
The network is trained until no further loss improvement is observed. Learning converges within 36 hours on a single GPU.

\begin{figure*}[t]
\newcommand{\kpimw}{1.2cm}
\newcommand{\textraise}{0.6cm}
\newcommand{\textraisew}{0.6cm}
\newcommand{\hortabsep}{0.3cm}

\small
\centering

{\tablinesep=0.1pt\tabcolsep=0.3pt

\begin{tabular}{p{1.6cm}cccc}
\raisebox{\textraise}{\pbox{1.3cm}{confidence\\$\sigma^{-1}$}} &
\includegraphics[width=\kpimw]{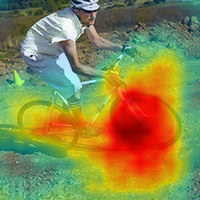} &
\includegraphics[width=\kpimw]{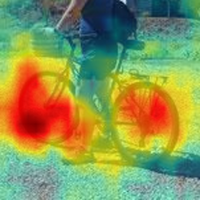} &
\includegraphics[width=\kpimw]{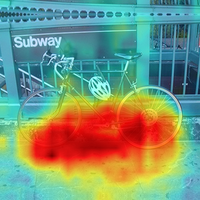} &
\includegraphics[width=\kpimw]{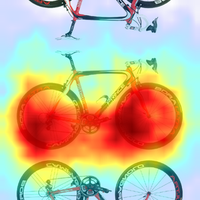}
\\
\raisebox{\textraise}{\pbox{1.3cm}{f. channel\\$[\phi(\bx)]_c$}} & 
\includegraphics[width=\kpimw]{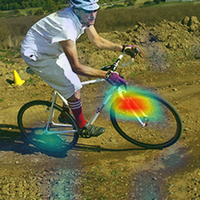} &
\includegraphics[width=\kpimw]{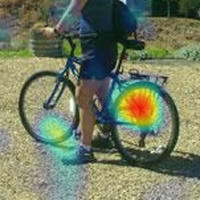} &
\includegraphics[width=\kpimw]{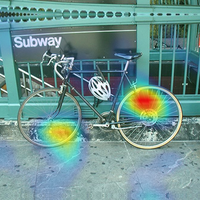} &
\includegraphics[width=\kpimw]{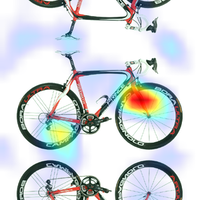}
\\
\raisebox{\textraise}{\pbox{1.3cm}{f. channel\\$[\phi(\bx)]_c$}} & 
\includegraphics[width=\kpimw]{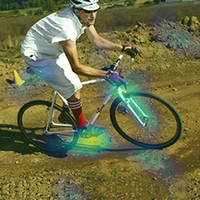} &
\includegraphics[width=\kpimw]{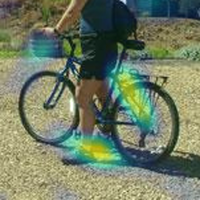} &
\includegraphics[width=\kpimw]{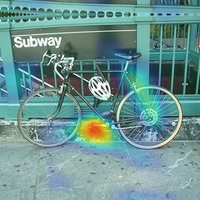} &
\includegraphics[width=\kpimw]{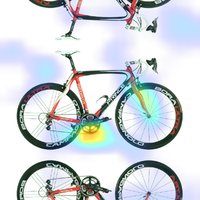}
\end{tabular}\hspace{\hortabsep}
\begin{tabular}{cccc}
\includegraphics[width=\kpimw]{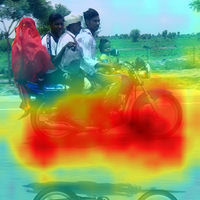} &
\includegraphics[width=\kpimw]{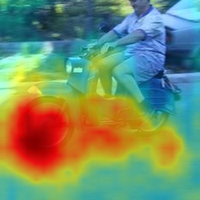} &
\includegraphics[width=\kpimw]{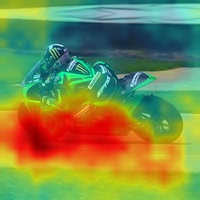} &
\includegraphics[width=\kpimw]{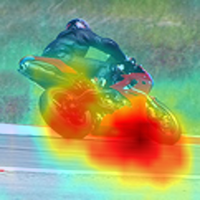}
\\
\includegraphics[width=\kpimw]{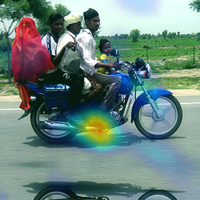} &
\includegraphics[width=\kpimw]{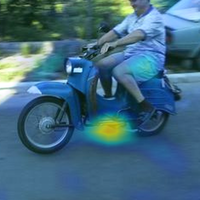} &
\includegraphics[width=\kpimw]{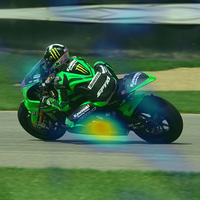} &
\includegraphics[width=\kpimw]{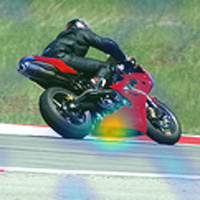} 
\\
\includegraphics[width=\kpimw]{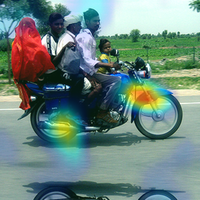} &
\includegraphics[width=\kpimw]{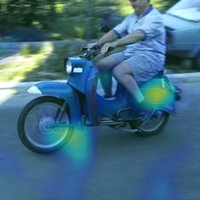} &
\includegraphics[width=\kpimw]{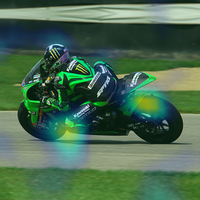} & 
\includegraphics[width=\kpimw]{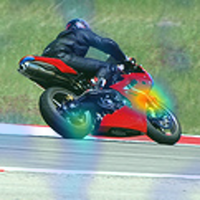}
\end{tabular}\hspace{\hortabsep}
\begin{tabular}{cccc}
\includegraphics[width=\kpimw]{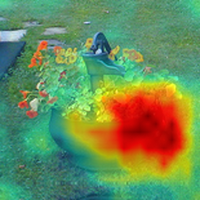} &
\includegraphics[width=\kpimw]{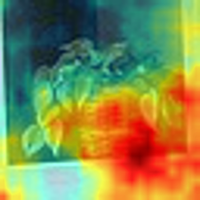} &
\includegraphics[width=\kpimw]{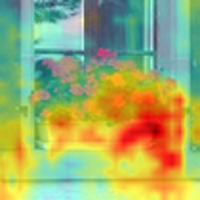} &
\includegraphics[width=\kpimw]{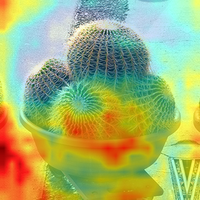}
\\
\includegraphics[width=\kpimw]{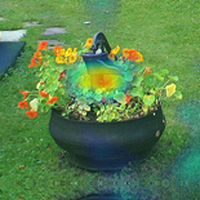} &
\includegraphics[width=\kpimw]{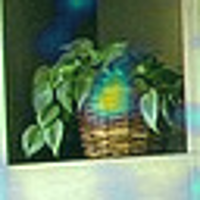} &
\includegraphics[width=\kpimw]{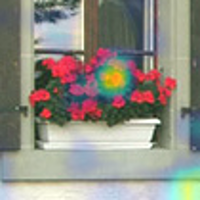} &
\includegraphics[width=\kpimw]{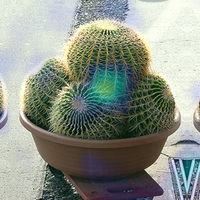}
\\
\includegraphics[width=\kpimw]{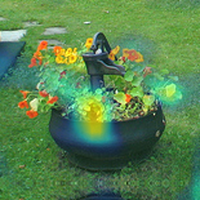} &
\includegraphics[width=\kpimw]{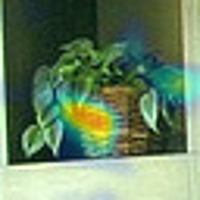} &
\includegraphics[width=\kpimw]{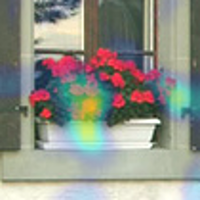} &
\includegraphics[width=\kpimw]{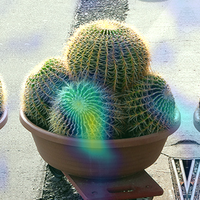}
\end{tabular}


\begin{tabular}{p{1.6cm}cccc}
\raisebox{\textraise}{\pbox{1.3cm}{confidence\\$\sigma^{-1}$}} &
\includegraphics[width=\kpimw]{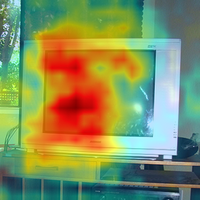} &
\includegraphics[width=\kpimw]{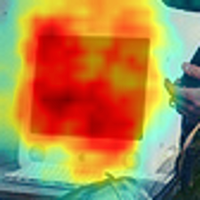} &
\includegraphics[width=\kpimw]{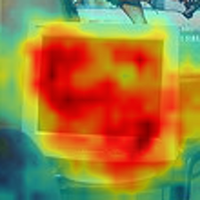} &
\includegraphics[width=\kpimw]{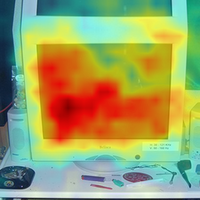}
\\
\raisebox{\textraise}{\pbox{1.3cm}{f. channel\\$[\phi(\bx)]_c$}} & 
\includegraphics[width=\kpimw]{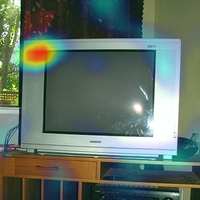} &
\includegraphics[width=\kpimw]{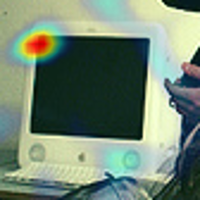} &
\includegraphics[width=\kpimw]{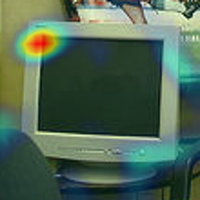} &
\includegraphics[width=\kpimw]{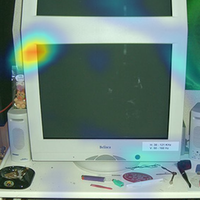}
\\
\raisebox{\textraise}{\pbox{1.3cm}{f. channel\\$[\phi(\bx)]_c$}} &
\includegraphics[width=\kpimw]{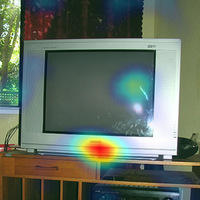} &
\includegraphics[width=\kpimw]{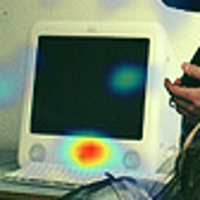} &
\includegraphics[width=\kpimw]{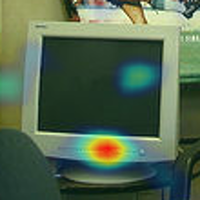} &
\includegraphics[width=\kpimw]{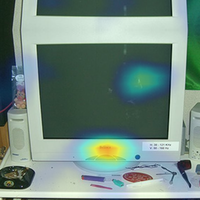}
\end{tabular}\hspace{\hortabsep}
\begin{tabular}{cccc}
\includegraphics[width=\kpimw]{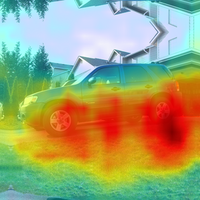} &
\includegraphics[width=\kpimw]{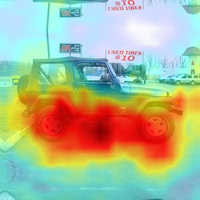} &
\includegraphics[width=\kpimw]{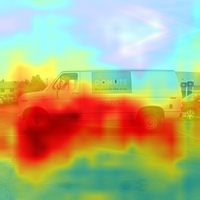} &
\includegraphics[width=\kpimw]{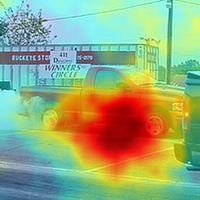} 
\\
\includegraphics[width=\kpimw]{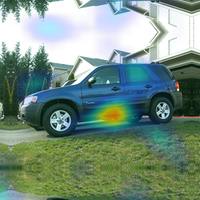} &
\includegraphics[width=\kpimw]{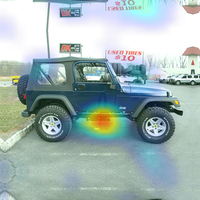} &
\includegraphics[width=\kpimw]{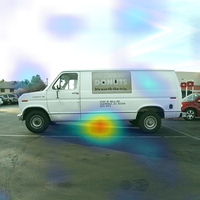} &
\includegraphics[width=\kpimw]{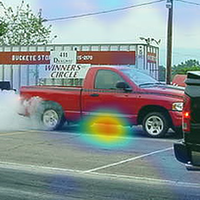}
\\
\includegraphics[width=\kpimw]{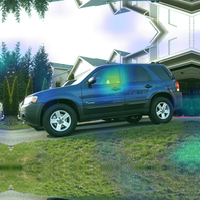} &
\includegraphics[width=\kpimw]{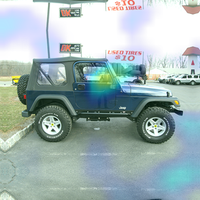} &
\includegraphics[width=\kpimw]{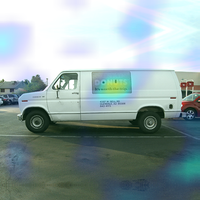} &
\includegraphics[width=\kpimw]{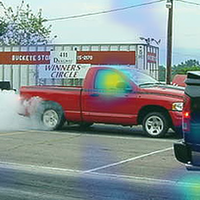}
\end{tabular}\hspace{\hortabsep}
\begin{tabular}{cccc}
\includegraphics[width=\kpimw]{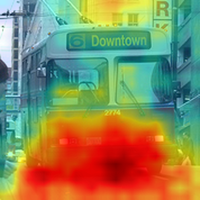} &
\includegraphics[width=\kpimw]{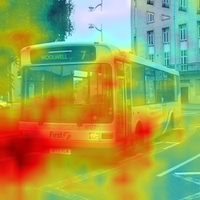} &
\includegraphics[width=\kpimw]{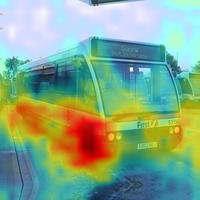} &
\includegraphics[width=\kpimw]{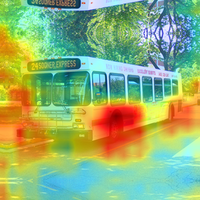}
\\
\includegraphics[width=\kpimw]{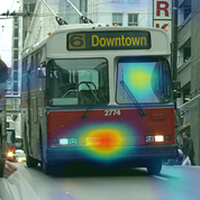} &
\includegraphics[width=\kpimw]{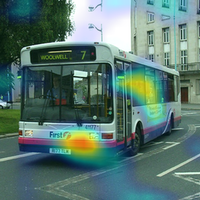} &
\includegraphics[width=\kpimw]{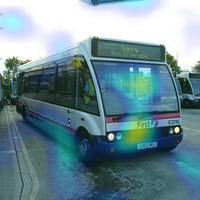} &
\includegraphics[width=\kpimw]{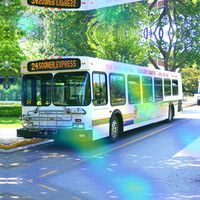}
\\
\includegraphics[width=\kpimw]{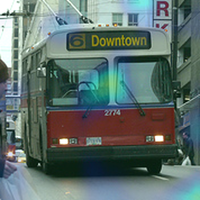} &
\includegraphics[width=\kpimw]{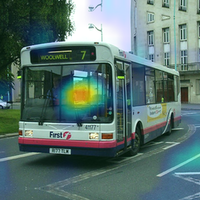} &
\includegraphics[width=\kpimw]{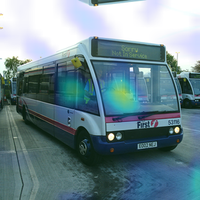} &
\includegraphics[width=\kpimw]{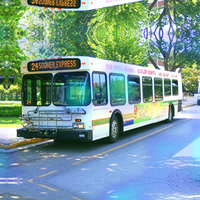}
\end{tabular}

}

\caption{\textbf{Qualitative analysis of the learned equivariant feature representation $\phi$} 
visualizing predicted confidence maps $\sigma^{-1}$ and several responses $\max([\phi(\bx)]_c,0)$ of different channels $c$ of the representation, for six different categories.}\label{f:confidence-kp}
\end{figure*}

\section{Experiments}\label{s:exp}

We first show qualitative results of our self-learning approach (\cref{sec:exp_qualitative}). Then, we quantitatively evaluate for the semantic matching (\cref{sec:exp_matching}) and for the keypoint detection (\cref{sec:exp_keypoint}) tasks.

\subsection{Qualitative analysis} \label{sec:exp_qualitative}

We first qualitatively analyze the nature of the learned feature representation. \Cref{f:confidence-kp} considers six categories and shows, for four images of each category, the confidence maps $\sigma(\bx)^{-1}$ along with example rectified responses $\max([\phi(\bx)_u]_c,0)$ for several feature channels $c$ of the learned representation.  It can be observed that the responses resemble distinct keypoint detectors that fire consistently across different instances of a category, even in the presence of large intra-class variations. Furthermore, the confidence predictor $\sigma(\bx)^{-1}$ can be interpreted as a generic detector of distinct areas of the image foreground.

\subsection{Semantic matching} \label{sec:exp_matching}

\begin{figure*}[h!]  
  \includegraphics [width=0.5\linewidth] {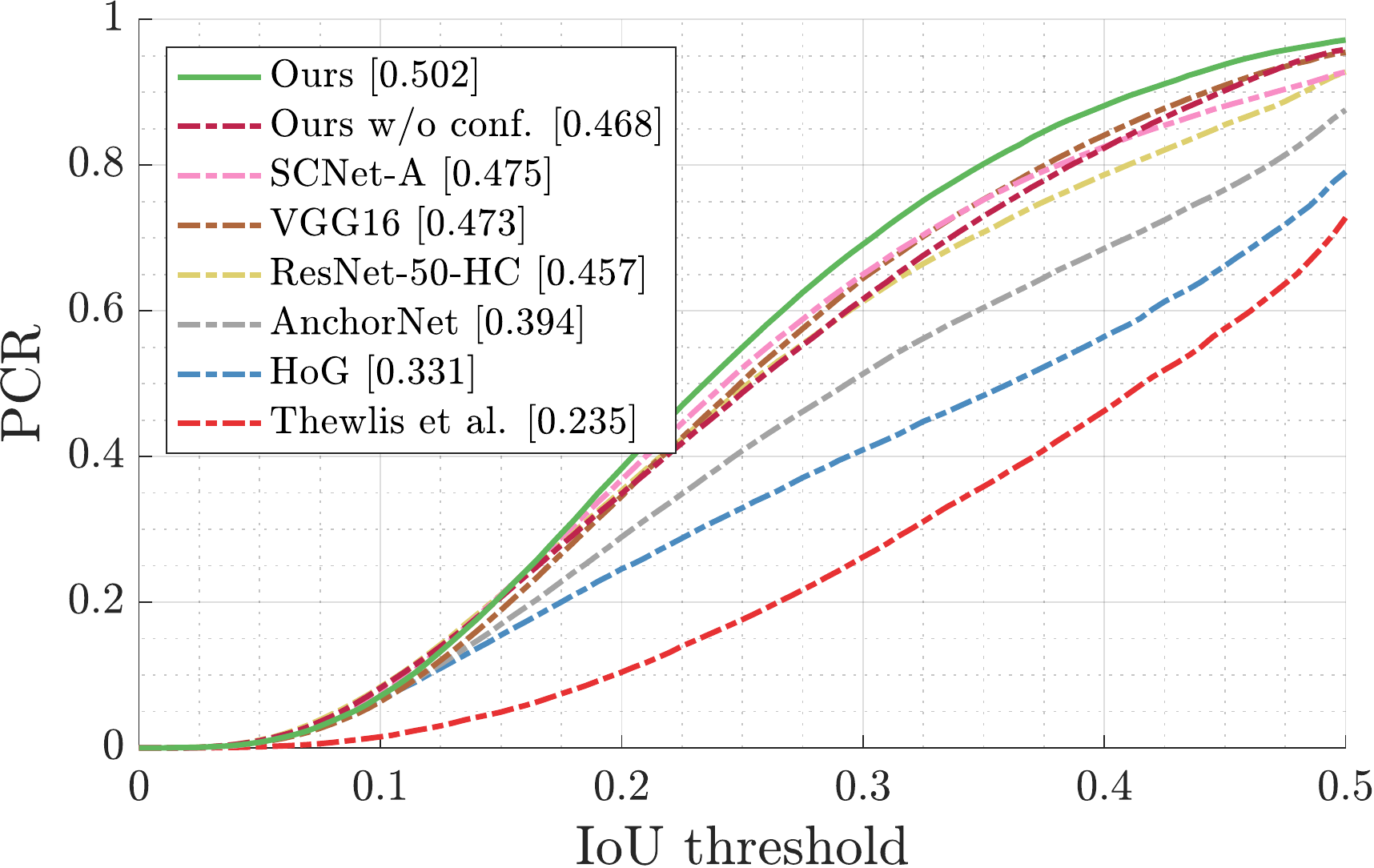}
  \includegraphics [width=0.5\linewidth] {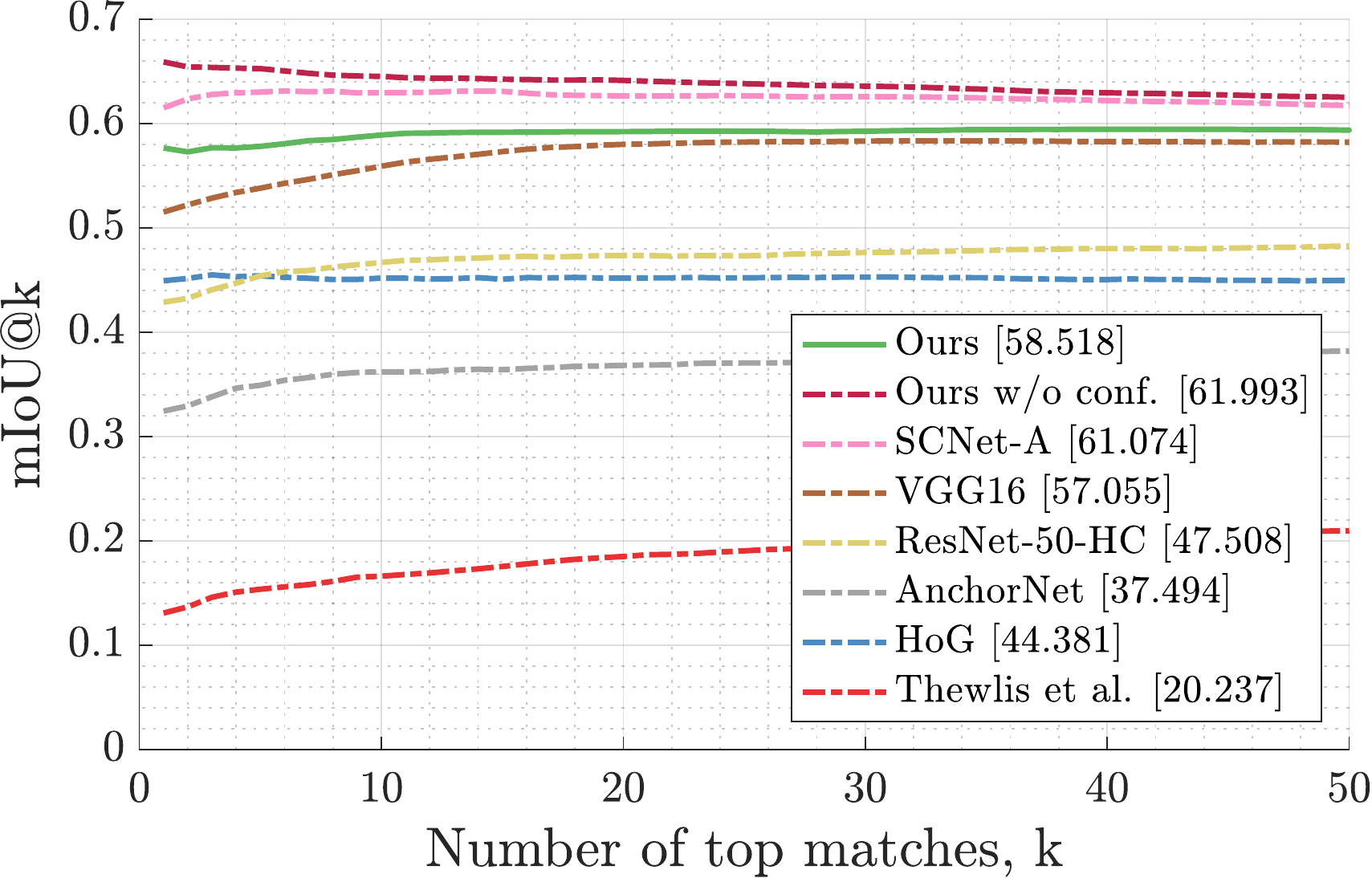}
  \label{fig:Ng2}
\vspace{-0.3cm}
\caption{\label{f:pcr}\textbf{Region matching performance on PF-Pascal}. Features are matched directly without any spatial regularization.
Left: region matching precision (PCR). Right: region matching accuracy (mIoU@k). Note that unlike all other reported approaches, SCNet-A \cite{han2017scnet} is a fully supervised method. }
\end{figure*}

We first assess our method on the problem of semantic matching and compare it to other unsupervised and weakly-supervised approaches for learning geometry-aware representation. In particular, we follow the dataset and experimental protocol of \cite{ham2016} and consider the problem of establishing correspondences between bounding box proposals and keypoints extracted in pairs of images.
 
\myparagraph{Compared approaches}

We compare our learned dense features to five existing feature representations. First, in order to demonstrate the improvement of our self-learning approach over the pre-trained (using only image-level labels)  ResNet-50 model, we consider \textbf{ResNet-50-HC} which is a hypercolumn architecture that pools features from the res3c, res4c, res5c layers and separately upsamples them to a common spatial size. 
In order to demonstrate the benefits of the probabilistic introspection, we also present results of \textbf{Ours w/o conf.} which is our method trained by optimizing the non-probabilistic loss function from \cref{e:matching}.
Then, to provide a direct comparison with approaches that tackle the geometric feature learning task, we report the results of \cite{novotny17anchornet} and \cite{thewlis17unsupervised}. For \textbf{AnchorNet}~\cite{novotny17anchornet}, we use their public class-agnostic model. To provide a fair comparison with the method of \textbf{Thewlis \etal}~\cite{thewlis17unsupervised}, we train their method on the  same dataset as used for our features. To establish a baseline, we explore three variants of the base architecture proposed in \cite{thewlis17unsupervised}:  a model with $10$ landmarks (as proposed in the original work), a model with $64$ landmarks (to increase model capacity)  and finally a modified, class specific architecture which learns a set of $64$ landmarks \textit{per-class}. In practice, we found the second design to be most effective, and
therefore, all reported results use this option.\footnote{While this approach has been shown to be effective under more constrained conditions, we were unable to achieve robust learning dynamics when applying it to our task.}
The last baseline uses pool4 features from the \textbf{VGG16} architecture \cite{simonyan2014very} pre-trained on the ImageNet image classification task. We selected these features, since they are the basis of current state-of-the-art semantic matching approaches \cite{rocco17convolutional,han2017scnet,ham2016}. Alongside other unsupervised and weakly supervised methods, we also compare against the fully supervised \mbox{\textbf{SCNet-A}} architecture introduced in~\cite{han2017scnet}.

For our approach, matching descriptors are produced by exploiting the confidence prediction capacity of our model, scaling the outputs of the final layer by the inverse of the predicted uncertainty $\sigma$. We then follow the simple approach developed in \cite{han2017scnet}, by applying ROI-pooling with bin size $7 \times 7$ to each proposal region resulting in a feature vector comprising these scaled representations. We further pool and concatenate res4c features from a lower layer of our network. In order to produce a dense warping field for keypoint matching we employ the sd-filtering as done in \cite{ham2016,han2017scnet}.
For keypoint matching, following other approaches \cite{ham2016,rocco17convolutional,han2017scnet}, we modify our original ResNet50-based architecture by replacing the network trunk with the VGG16 architecture truncated after the pool4 features and terminated as described in \cref{s:details}. This network was trained on all 20 PASCAL classes of the ImageNet dataset according to the same learning schedule as described in \cref{s:details}. For this architecture, instead of res4c features we pool and concatenate the pool4 features.

Since our objective is to assess \textit{feature quality}, we evaluate each method without using any spatial regularization (such as \eg Local Offset Matching~\cite{ham2016}, joint warp estimation~\cite{rocco17convolutional}, or MRFs with geometric potentials \cite{ufer2017deep}).\footnote{The development of effective spatial regularization methods forms an important, but orthogonal line of research to the focus of our work.}

\myparagraph{Dataset}
We evaluate our approach on the PF-PASCAL dataset \cite{ham2016} which contains pairs of images which have been fully annotated with keypoints for $20$ object classes. Each method is evaluated with a set of $1000$ object proposals per image, generated with the Randomized Prim (RP) method~\cite{manen2013prime}. Following~\cite{han2017scnet}, performance is reported on the \textit{test} partition, which comprises $302$ image pairs.

\begin{table}[t]
\centering
\begin{tabular}{lr|lr}
  \hline
\textbf{Method}                              & \textbf{PCK} & \textbf{Method}               & \textbf{PCK} \\ \hline
Thewlis et al. \cite{thewlis17unsupervised}  & 14.4         & ResNet50-HC \cite{he2015deep} & 64.0 \\
AnchorNet  \cite{novotny17anchornet}         & 56.3         & SCNet-A \cite{han2017scnet}   & 66.3 \\
VGG16 \cite{ham2016}                         & 62.3         & Ours w/o conf.                & 60.6 \\
gCNN \cite{rocco17convolutional}             & 62.6         & \textbf{Ours}                 & 66.5 \\
\hline
\end{tabular} \vspace{0.1cm}
\caption{\textbf{Keypoint matching performance on PF-Pascal} reporting PCK@0.1 for our method and existing approaches.}
\label{t:pck}
\end{table}

\myparagraph{Evaluation}
We report results under the standard PCR (probability of correct regions) and mIoU@$k$ (mean intersection over union of the best $k$ matches) metrics introduced in \cite{ham2016}. PCR aims to capture the accuracy of overall assignment, while mIoU@$k$ reflects the reliability of matching scores. 
Following the common practice on this dataset, keypoint matching is assessed by reporting PCK@$\alpha$ with the misalignment sensitivity threshold $\alpha$ set to 0.1. All evaluations are conducted using the public implementation provided by the authors of~\cite{han2017scnet}. 

\myparagraph{Results}
The region matching results are shown in \Cref{f:pcr}. First, we observe that our approach significantly outperforms previous representations trained with a comparable amount of supervision: AnchorNet~\cite{novotny17anchornet}, the method of Thewlis \etal~\cite{thewlis17unsupervised}, and VGG16 \cite{simonyan2014very}. Second, we see that, interestingly, our self-supervised features perform on par with the model SCNet-A of~\cite{han2017scnet} which is in fact \textit{fully supervised} with keypoint annotations. These observations are encouraging also due to the fact that our representation was
trained only on rigid classes while the PF-Pascal dataset also contains a large portion of the non-rigid ones.

Results for keypoint matching are present in \Cref{t:pck}. Similar to region matching, we observe improvements over other approaches trained with comparable level of supervision. Furthermore, our results are again on par with the fully supervised SCNet-A~\cite{han2017scnet}.
We observe a decrease in matching performance with Ours w/o conf. which validates the importance of the proposed instrospection mechanism.

\begin{figure}[t!]
\includegraphics [width=\linewidth] {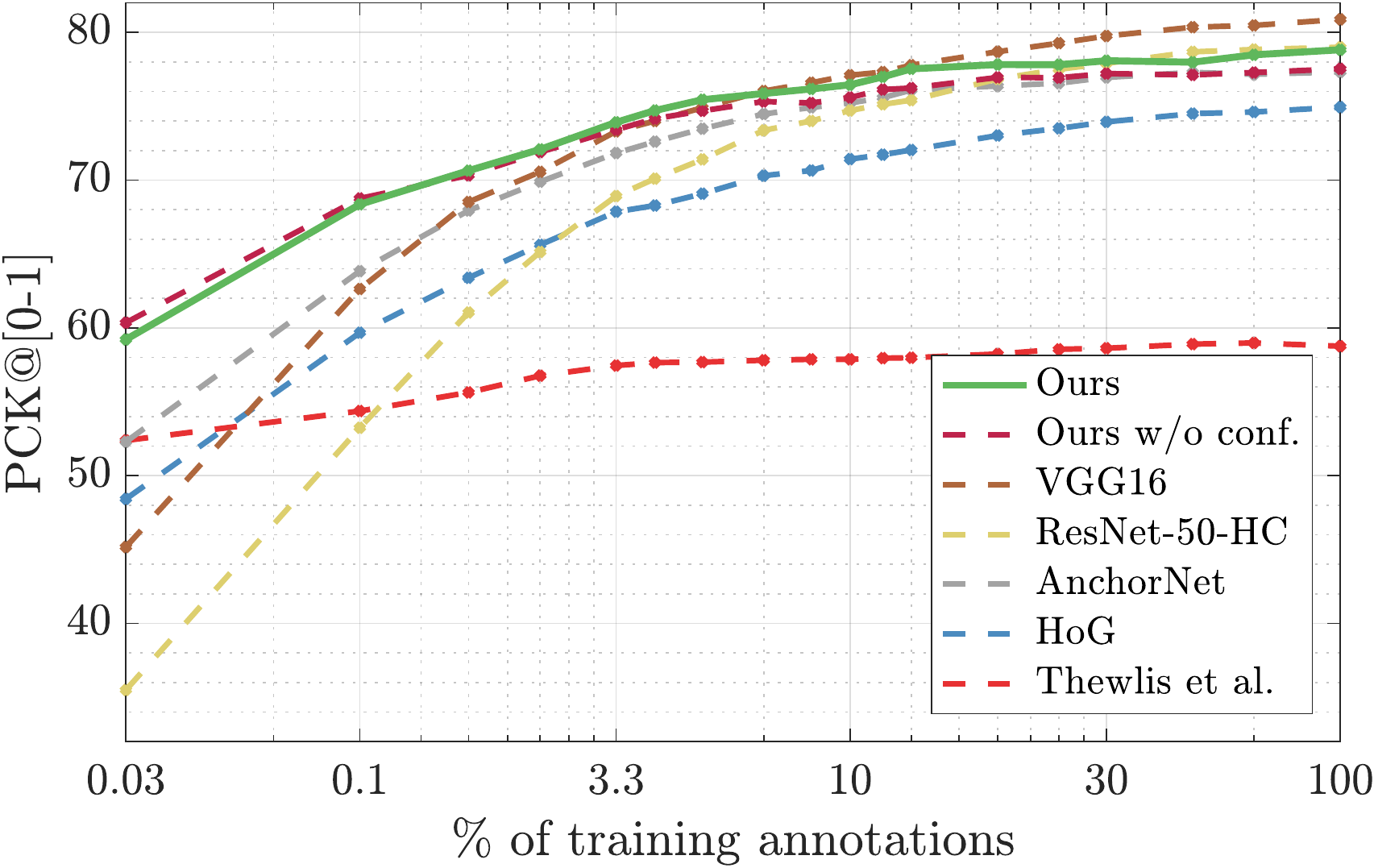}
\caption{\textbf{Keypoint prediction on Pascal3D}. We report the area under the PCK-over-alpha curve as a function of the  number of training annotations.} 
\label{f:pckVSnAnno}
\end{figure} 

\subsection{Few-shot keypoint detection} \label{sec:exp_keypoint}

In \cref{sec:exp_qualitative} we have observed that the learned features often correspond to distinctive object parts. Those do not necessarily have a semantic meaning, as demonstrated in \cite{thewlis17unsupervised}, but they can still be used as anchors that facilitate the detection of semantic parts.  Following \cite{thewlis17unsupervised}, in this section we tackle the task of semantic keypoint detection where our learned representation as well as competitors is used as input features for a keypoint predictor. The keypoint detection performance then serves as an estimate of how well the respective representations encode the geometrical structure of visual categories. We depart from \cite{thewlis17unsupervised} and we consider a significantly more challenging setting with out-of-plane rotations and large appearance variations.

Furthermore, an important feature of successful geometric representations is how well they facilitate transfer of information from a very limited number of annotated samples. Hence, here we consider keypoint detection with few-shot supervision where a training set of object keypoint annotations is gradually extended with new training samples while monitoring the performance on a held-out test set.

\myparagraph{Dataset} We use the keypoint annotations from the original Pascal3D dataset~\cite{xiang2014beyond}. The few-shot keypoint predictors are trained on the ``train'' set of Pascal3D and evaluated on the held-out ``val'' set. Following common practice \cite{tulsiani2015viewpoints}, knowledge of a ground truth bounding box as well as the depicted object's class is assumed during both training and testing. The task is evaluated using the probability of correct keypoint measure (PCK) introduced in~\cite{yang2013articulated}. A keypoint prediction is regarded as correct if  its distance from the corresponding ground truth annotation is lower than $\alpha\times\max\{w,h\}$,  where $w,h$ are the object bounding box dimensions and $\alpha$ controls the sensitivity of the measure to misalignments. For each class, PCK corresponds to the ratio between the number of correct predictions and the total number of keypoint annotations. Similar to the PCR metric, we integrate the measure over all possible $\alpha$ values  and report the average over the 12 Pascal3D object classes. 

\myparagraph{Keypoint predictor} Our keypoint predictor consists of a 512-channel $3\times 3$ convolutional layer with stride 1 followed by batch normalization, ReLU and a final $3\times 3$ convolutional layer with stride 1 terminated by the sigmoid activation function.  Each channel of the final layer then serves as a response map of the corresponding keypoint class. The loss minimizes the weighted $\ell_2$ distance between the ground truth heatmap and the corresponding prediction as proposed in \cite{tulsiani2015viewpoints}. The evaluation process alternates between training the keypoint detector, evaluating its performance and adding a new set of training annotations consisting of an equal number of randomly sampled images per class.  For each round, the detector is trained for 3 epochs making sure that at least 500 training steps are performed for each epoch. Detector parameters are initialized with the model from the previous round. The experiment is run three times with different random seeds and we report an average over PCKs. 

\myparagraph{Results} Results of the few-shot detection experiments are reported in \Cref{f:pckVSnAnno}.  Our method surpasses all the compared approaches when a small percentage of the training annotations is available, and in particular the methods of \cite{novotny17anchornet}, \cite{thewlis17unsupervised}, and \cite{rocco17convolutional}, while performing on par with the best competitor on this task (VGG16~\cite{simonyan2014very}) when the full training set is used. Similar to the semantic matching experiments \cref{sec:exp_matching}, we observe significant drop in performance of the method from \cite{thewlis17unsupervised}.
Ours w/o conf. obtains similar results to the proposed method. This is likely due to the fact that the detection dataset does not contain a large quantity of background clutter because the evaluated instances are always cropped using a tight ground truth bounding box.
\section{Conclusions}\label{s:conc}

In this paper, we have presented a self-supervised method that can pre-train features useful to reason about the geometry of object categories in tasks such as part localization and semantic matching. The method combines the robustness of recent approaches such as AnchorNet with the geometric prior induced by invariance to synthetic image transformations. This allows to train features that excel at these geometric tasks using only images with class-level annotations. We have shown that these features outperform all other pre-training methods in semantic matching and part localization. In the case of the first task, our features perform on par with a fully-supervised approach.

\paragraph{Acknowledgments.} The authors gratefully acknowledge the support of EPSRC AIMS, Seebibyte and ERC 677195-IDIU. The authors would also like to thank James Thewlis for kindly sharing code.

\clearpage
\appendix
\begin{strip}%
 \centering
 \Large
 \textbf{
  Self-supervised Learning of Geometrically Stable Features Through\\Probabilistic Introspection \\ \vspace{0.3cm} \textit{Appendix}
 }
\end{strip}

In the supplementary material below, we present an ablation study of the components of our method (\cref{s:ablation}).  
In \cref{s:weaksup}, we also provide details of the weakly supervised method that produced the bounding box annotations used to train our model. 

\section{Ablation studies \label{s:ablation}}

In addition to the results reported in sections 4.2. and 4.3. of the paper, we report additional ablation experiments that validate the contribution of the proposed components of our method.

In order to show the improvements over the base architecture that was used to initialize our network, we also compare against the res5c features from the version of the pretrained ResNet-50 model, the filters of which were dilated as explained in section 3.6. in the paper (\textbf{ResNet-50-dilated}). 

Furthermore, to provide an extended comparison with alternative matching loss formulations, a flavour of our method, abbreviated as \textbf{Contrastive}, implements the contrastive loss formulation from \cite{choy2016universal}.

We also test three more methods that assess the sensitivity of the proposed approach to the utilized dataset. We include results for our method trained with ground truth bounding box labels (\textbf{Ours-GTbox}), rather than the weakly supervised detections used in the original formulation, to enable an assessment of the method's robustness to the usage of imperfect bounding box annotations. Another variation of our method, \textbf{Ours-NObox}, does not use any bounding box annotations. Finally, \textbf{Ours-nonrigid} uses all 20 PASCAL categories for training as opposed to the original training setup that used images of the 12 rigid classes.

\begin{figure}[ht!]  
  \includegraphics [width=0.95\linewidth] {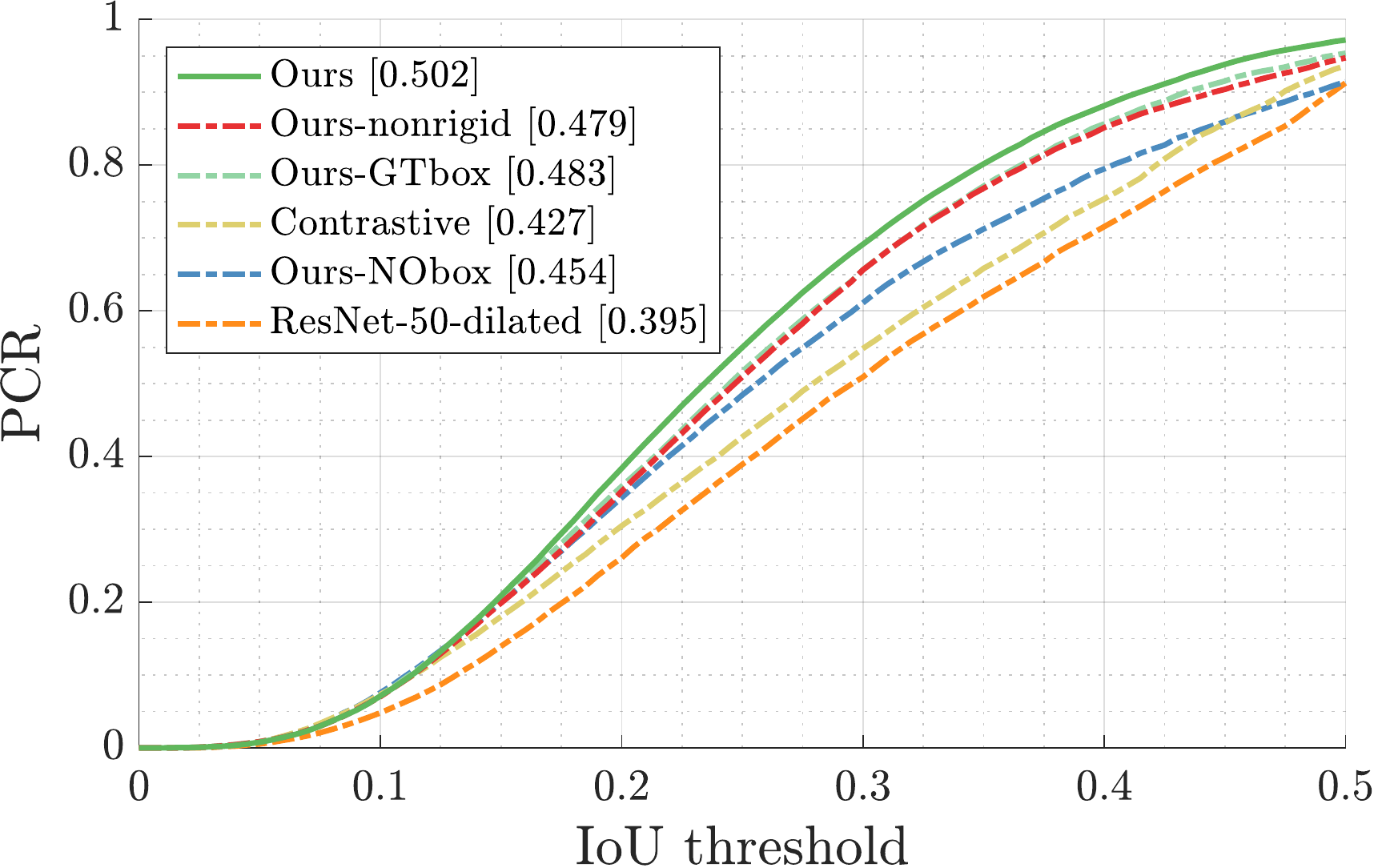}
\caption{\label{f:sup-sem-match}\textbf{Ablation study on PF-Pascal}. The region matching performance of several variants of our method (see \cref{s:ablation} for details of each variant).}
\end{figure}

\begin{figure}[ht!]
\includegraphics [width=\linewidth] {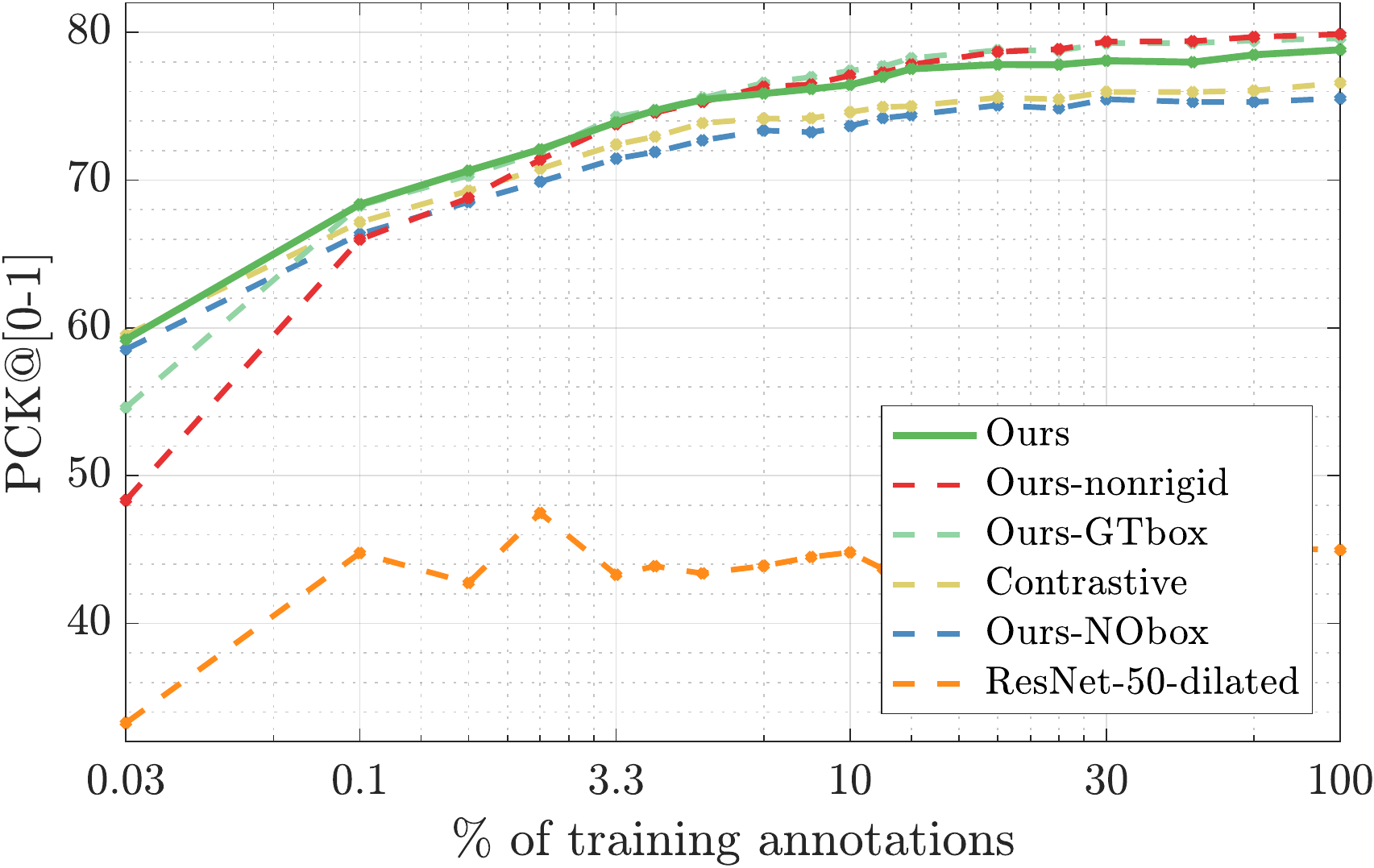}
\caption{\textbf{Ablation study on the few-shot keypoint detection task on Pascal3D}. We report the area under the PCK-over-alpha curve as a function of the  number of training annotations for several variants of our method. For details of each variant see \cref{s:ablation}.}
\label{f:sup-keypoint-det}
\end{figure}

\begin{figure*}[ht!]
\includegraphics [width=\linewidth] {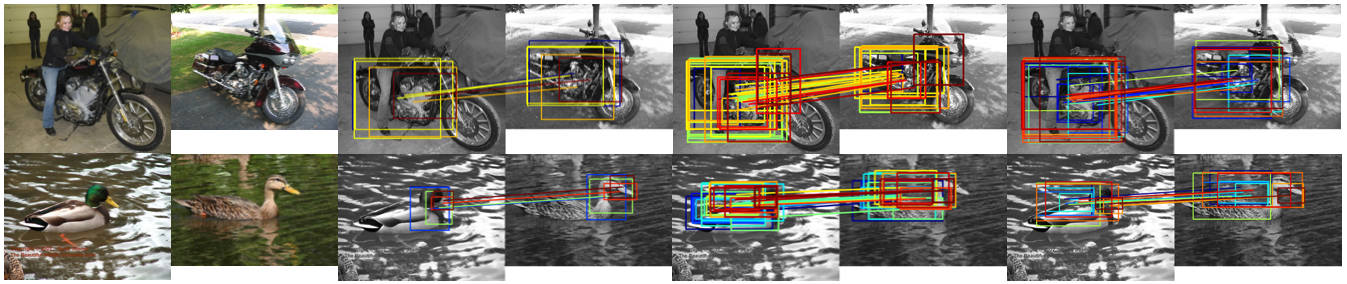}
\caption{{\bf Region matching examples} for pairs of motorbike (top) and duck (bottom) images.  From left to right: source and target images, HOG with NAM matching~\cite{ham2016}, ours, SCNet-A~\cite{han2017scnet}. We show correctly matched boxes, color-coded according to matching score (red: higher, blue: lower).}\label{f:pf-warps}
\end{figure*}

All variants were evaluated on both the semantic matching and keypoint prediction tasks.
The results of the semantic matching experiments are reported in \cref{f:sup-sem-match}
while \cref{f:sup-keypoint-det} contains the results of the few-shot keypoint prediction task.

The results indicate that for both semantic matching and keypoint detection the performance of the ground-truth supervised setup is on par with the proposed weakly supervised setup. This shows that, with the inclusion of the probabilistic introspection mechanism, the method has good robustness to annotation noise.  
The performance of our method trained with the non-rigid categories is on par with the rigid case for proposal matching. We observe a decrease in performance for the keypoint detection task. This is because the few-shot detection dataset consists of only rigid classes and adding the non-rigid ones to the training set makes the features less specialized for the final task. The variant which trains features via the contrastive loss gives lower performance.

\subsection{ Keypoint detection - detector validation }

In section 4.3. in the paper, we reported results for a 
keypoint detector with a design closely related to that of \cite{tulsiani2015viewpoints}. 
In order to validate the implementation of the detector, we provide
a comparison against the results of the fully supervised detector from \cite{tulsiani2015viewpoints}.
When using all available annotations and the Resnet-50-HC descriptors, the mean PCK ($\alpha=0.1$) 
over the 12 rigid classes of the Pascal3D test set is 54.7. This is on par with the best single-model
result from \cite{tulsiani2015viewpoints} (53.3 PCK), validating our keypoint predictor
as a representative proxy for evaluating the quality of our feature baselines.

\section{Weakly supervised detections \label{s:weaksup}}

Here we give details of the weakly supervised detector used to provide bounding box annotations for our method, as discussed in Sec. 3.6 of the paper.  We use the \texttt{vgg-f}-based model described in \cite{bilen16}, which is trained using EdgeBox proposals\cite{zitnick2014edge} and the image-level labels of the Pascal VOC $2007$ detection dataset \cite{everingham2010pascal}. To produce bounding box predictions for the ImageNet dataset, we follow the multiscale evaluation technique described in \cite{bilen16}, averaging predictions over five scales and flipped copies of each scale.  To form our training set, we then select top scoring box for each class label present in the image.  In order to maintain a high quality of box annotation, we do not include boxes whose scores fall below the median detector score of the given class (the median is computed after filtering scores which fall below the noise score threshold of $0.001$ given in the public implementation\footnote{\url{https://github.com/hbilen/WSDDN}} of \cite{bilen16}).

\section{Qualitative results \label{s:qualitative}}

Additional qualitative results for the semantic matching task ok PF-Pascal are present in \cref{f:pf-warps}. We show the matching regions for two example pairs, for the method of \cite{ham2016}, ours, and the fully-supervised method of SCNet-A.

{\small\bibliographystyle{ieee}\bibliography{refs}}

\end{document}